\documentclass[10pt,twocolumn,letterpaper]{article}

\usepackage{cvpr}              

\definecolor{cvprblue}{rgb}{0.21,0.49,0.74}
\usepackage[pagebackref,breaklinks,colorlinks,allcolors=cvprblue]{hyperref}
\usepackage{epigraph}
\usepackage{csquotes}
\usepackage{newtxtext,newtxmath}
\usepackage{multirow}
\usepackage{comment}

\def\paperID{12334} 
\def\confName{CVPR}
\def\confYear{2026}

\title{Frequency Switching Mechanism for Parameter-Efficient Multi-Task Learning}

\author{
Shih-Wen Liu$^{1}$ \quad
Yen-Chang Chen$^{1}$ \quad
Wei-Ta Chu$^{1}$ \quad
Fu-En Yang$^{2}$ \quad
Yu-Chiang Frank Wang$^{2}$ \\
$^{1}$National Cheng Kung University, Tainan, Taiwan \\
$^{2}$NVIDIA Research, Taiwan \\
{\tt\small \{ne6134079, f74091174, wtchu\}@gs.ncku.edu.tw} \quad
{\tt\small \{fredy, frankwang\}@nvidia.com}
}

\begin{document}
\maketitle

\begin{abstract}
Multi-task learning (MTL) aims to enable a single model to solve multiple tasks efficiently; however, current parameter-efficient fine-tuning (PEFT) methods remain largely limited to single-task adaptation. 
We introduce \textbf{Free Sinewich}, a parameter-efficient multi-task learning framework that enables near-zero-cost weight modulation via frequency switching (\textbf{Free}). 
Specifically, a \textbf{Sine-AWB (Sinewich)} layer combines low-rank factors and convolutional priors into a single kernel, which is then modulated elementwise by a sinusoidal transformation to produce task-specialized weights.
A lightweight Clock Net is introduced to produce bounded frequencies that stabilize this modulation during training. 
Theoretically, sine modulation enhances the rank of low-rank adapters, while frequency separation decorrelates the weights of different tasks. 
On dense prediction benchmarks, Free Sinewich achieves state-of-the-art performance-efficiency trade-offs (e.g., up to +5.39\% improvement over single-task fine-tuning with only 6.53M trainable parameters), offering a compact and scalable paradigm based on frequency-based parameter sharing. 
Project page:
\href{https://casperliuliuliu.github.io/projects/Free-Sinewich/}{https://casperliuliuliu.github.io/projects/Free-Sinewich}.
\end{abstract}
\section{Introduction}
\epigraph{\itshape
  ''Match the frequency of the reality you want and you cannot help but get that reality.''
}{\textsc{Albert Einstein}}

\begin{figure}[t]
    \centering
    \includegraphics[width=\linewidth]{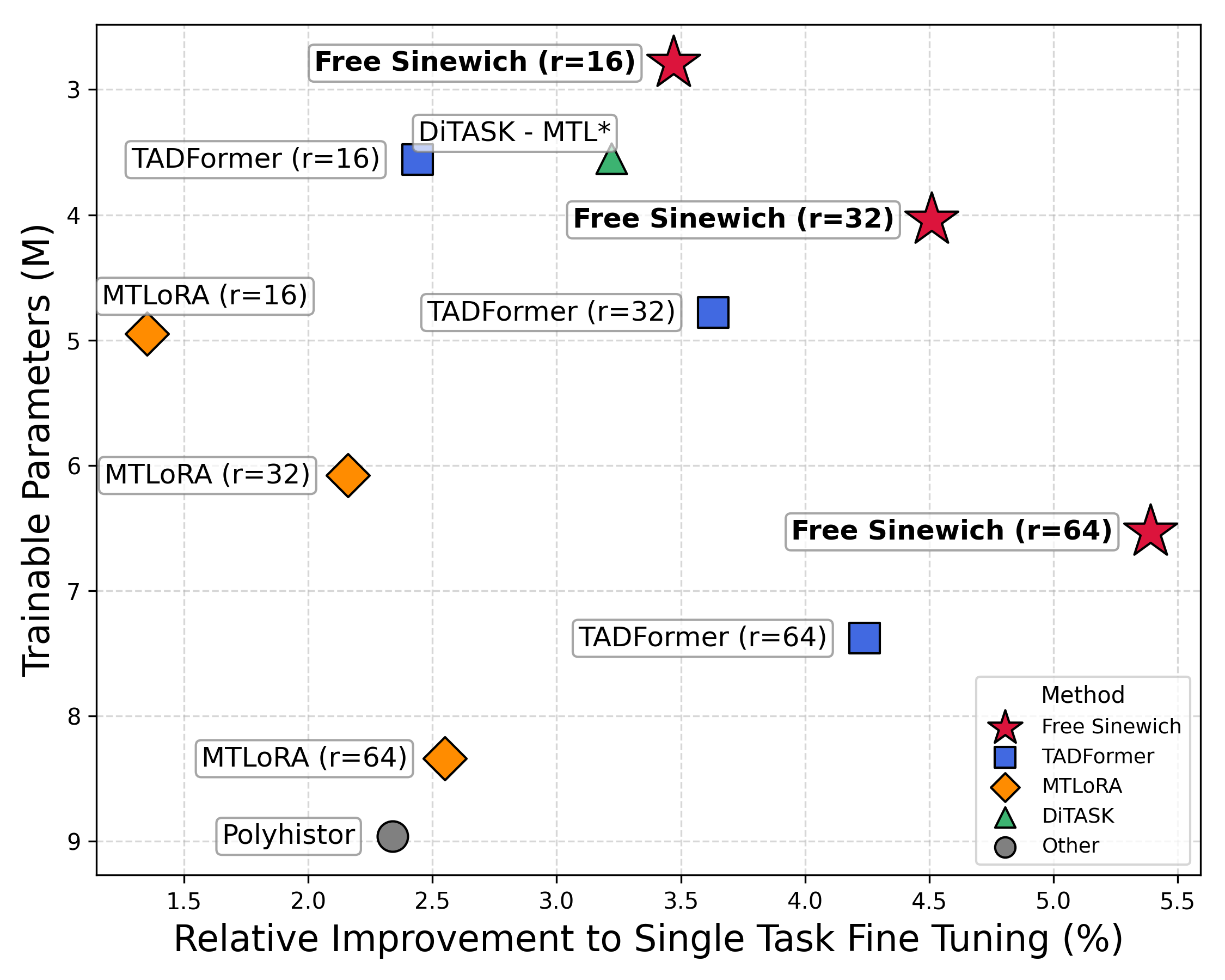}
    \caption{Overall comparison of different PEFT-MTL methods.
    $r$ denotes the rank for the low-rank decomposition modules.}
    \label{fig:comparison}
\end{figure}

Fine-tuning large pretrained models on individual downstream tasks has proven highly effective in computer vision~\cite{transfer}.
However, many real-world applications require a single model to handle multiple tasks simultaneously.
When tasks are intrinsically related, a multi-task learning (MTL) model that shares parameters across multiple tasks would improve overall performance and generalization in principle~\cite{mtl_survey,MTL_CV_survey}.
However, MTL often suffers from \emph{task conflict} and \emph{negative transfer}~\cite{gradient_surgery,Negative_survey,ConsMTL}, where gradients from different objectives interfere and degrade each other. 
Consequently, many approaches introduce \emph{task-specific} modules that sidestep interference but compromise true parameter sharing. 
The key question thus is: \emph{how can one retain parameter efficiency while enabling shared weights to behave task-specifically?} 

To improve efficiency and scalability, recent studies explore PEFT-MTL frameworks~\cite{peft_llm,peft_large,peft_A2Z}.
Methods such as MTLoRA~\cite{agiza2024mtlora} incorporate both task-agnostic and task-specific low-rank adapters on top of a shared backbone to balance shared representations and specialization. 
DiTASK~\cite{mantri2025ditask} leverages a differentiable homomorphic transformation on singular values, preserving pretrained subspaces while minimally injecting task-specific differences. 
TADFormer~\cite{baek2025tadformer} combines parameter-efficient prompting with dynamic task filters to condition convolutional layers on input features for dense scenarios. 
Despite their success, these approaches share structure or feature representations rather than truly reusing the same \emph{parameter set} across tasks.
They rely on auxiliary adapters to route information through separate paths, resulting in distinct parameter sets for each task-specific module.
Without parameter reuse, models cannot fully leverage the common knowledge across tasks, often resulting in redundant computation and less generalization.


The human brain has an elegant and highly efficient mechanism for multi-task processing. 
Rather than allocating independent neural circuits for each task, neuroscience suggests that the \emph{thalamocortical system} enables selective communication through \emph{oscillatory multiplexing}~\cite{akam2014oscillatory,Thalamocortical}. 
The same neuronal populations perform distinct functions at different oscillatory frequencies, reusing biological \enquote{hardware} by switching frequency instead of forming new regions~\cite{kucewicz2024high,pagnotta2024multiplexed}. 
This insight inspires us: \emph{can deep networks similarly reuse the same weights, by switching their frequency response to yield task-specific functions?}

Motivated by this biological mechanism, we propose \textbf{Free Sinewich}, a PEFT-MTL framework that achieves effective parameter reuse through a \emph{frequency-switching mechanism}. 
We design a learnable sinusoidal modulation that augments a shared low-rank adapter without adding extra parameters, so that the performance of multiple tasks can be boosted while fine-tuning parameters can be kept efficient. 
Figure~\ref{fig:comparison} shows a comparison of recent PEFT-MTL methods.  
For dense prediction benchmarks, our approach attains state-of-the-art accuracy with significantly fewer trainable parameters.
Our contributions are as follows:
\begin{itemize}
  \item We propose a frequency-switching PEFT-MTL framework that \emph{mimics brain-like oscillatory reuse}: a shared parameter base is modulated by task-dependent sine transformation to yield specialized weights efficiently. 
  \item We achieve state-of-the-art results across multiple dense prediction tasks with minimal additional parameters. 
\end{itemize}

\section{Preliminary}
\label{sec:prelim}
Low-rank adaptation (LoRA)~\cite{hu2022lora} is the most widely used strategy to efficiently fine-tune large models. We build the proposed method by enhancing LoRA in the following. 

Let $M \in \mathbb{R}^{m \times n}$ denote a low-rank matrix used to update a pre-trained weight matrix: 
\begin{equation}
M = AB^{\!\top} = \sum_{i=1}^{r} \boldsymbol{a}_i \boldsymbol{b}_i^{\!\top},
\qquad
A \in \mathbb{R}^{m \times r}, \; B \in \mathbb{R}^{n \times r}, 
\end{equation}
where $r \ll \min(m, n)$ is the intrinsic rank. 

\subsection{Sine Transformation}
\label{subsec:sine}
\paragraph{Sine-LoRA.}
In conventional LoRA, the capacity of a low-rank matrix to convey information is significantly reduced, often leading to worse performance compared to its full-rank counterparts. The recent Sine-LoRA~\cite{ji2025sine} proposes to enhance the representational power of a low-rank matrix by introducing a nonlinear transformation: 
\begin{equation}
M = \sin(\omega \cdot AB^{\!\top}), 
\end{equation}
where $\omega$ is a frequency parameter. The key insight is that applying element-wise sine mapping to a low-rank matrix can significantly raise its effective rank without increasing the number of trainable parameters. This mapping thus preserves parameter efficiency while improving expressivity. 

\begin{figure}[t]
  \centering
  \begin{subfigure}[t]{\linewidth}
    \centering
    \includegraphics[width=6.5cm]{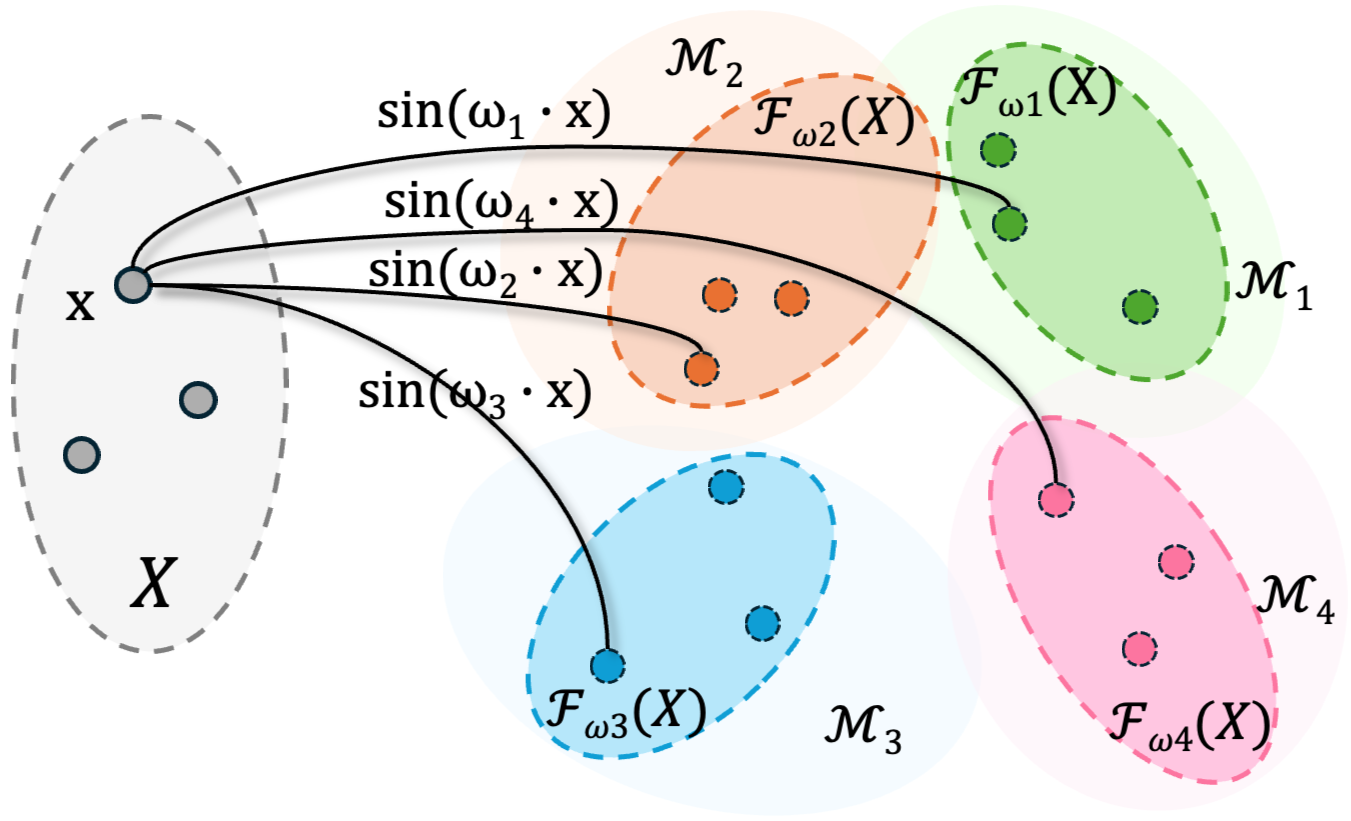}
    \caption{\textbf{Mapping between base space $X$ to task-specific space $\mathcal{M}_t$.}}
  \end{subfigure}
  \vspace{2mm}
  \begin{subfigure}[t]{\linewidth}
    \centering
    \includegraphics[width=6.5cm]{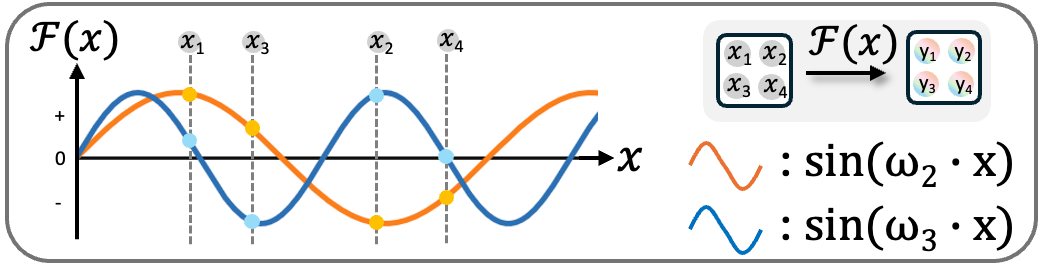}
    \caption{\textbf{Frequency-Switching through sine transformation.}}
  \end{subfigure}
  \vspace{-2mm}
  \caption{
Illustration of frequency switching.
(a) The base matrix space $X$ represents a shared parameter substrate reused across tasks.  
Each task-specific matrix space $\mathcal{M}_t$ denotes the domain where the transformed matrix resides.  
Different sine transformations with task-dependent frequencies $\omega_t$ map the same base matrix into $M_t$, producing task-specific matrices with distinct properties.  
(b) Each $\omega_t$ defines a unique sine wave, inducing a nonlinear mapping $\mathcal{F}_{\omega_t}$ that transforms the shared base matrix into task-specific representations elementwise.  
This illustrates how varying $\omega_t$ enables task specialization without parameter duplication.
}
  \label{fig:sine_map_full}
\end{figure}

\paragraph{Our formulation.}
We employ the same sine transformation but with a fundamentally different purpose.
Instead of boosting a single matrix, we construct a family of \emph{task-specific} matrices derived from one shared base $M_{base}$:
\begin{equation}
M_t = \mathcal{F}_{\omega_t}(M_{base}) = \sin(\omega_t \cdot AB^{\!\top}), \qquad t \in \mathcal{T}, 
\end{equation}
where $\mathcal{T}$ is the set of tasks (e.g., segmentation, saliency, normals), 
and $\omega_t$ is the task-specific frequency coefficient determined adaptively for different tasks.
As illustrated in Figure.~\ref{fig:sine_map_full}, each $\omega_t$ corresponds to a distinct sine wave, and therefore induces a different nonlinear mapping $\mathcal{F}_{\omega_t}$. 
This frequency-switching mechanism enables reusing a shared parameter base to achieve PEFT-MTL. 

\label{sec:prelim}

\subsection{AWB: Convolution-in-the-Middle for LoRA}
\label{subsec:awb_prelim}
For dense prediction tasks, spatial priors are crucial~\cite{khan2023survey}. 
Following TADFormer~\cite{baek2025tadformer} and ConvLoRA~\cite{zhong2024convlora}, we enhance the LoRA pathway by inserting a convolutional kernel inside the low-rank decomposition. 
Specifically, in addition to the LoRA factors $A \in \mathbb{R}^{m\times r}$ and $B \in \mathbb{R}^{n\times r}$, we introduce a convolution kernel $W \in \mathbb{R}^{r \times r \times K \times K}$ acting on the intermediate $r$ rank channels. 

\paragraph{Formulation.}
The forward path can be described as three steps: 
(i) a linear projection by the matrix $A$ that reduces the input channels from $m$ to $r$ channels;
(ii) a spatial convolution $W$ operating the $r$ channels;
(iii) a linear projection by the matrix $B^{\!\top}$ that expands the $r$ channels back to $n$.
Symbolically:
\[
X \;\xrightarrow{\;A\;} X_A
\;\xrightarrow{\;W\;} X_W
\;\xrightarrow{\;B^\top\;} Y
\]
where $X$ and $Y$ are the input and the output, respectively. 

Because convolution is a \emph{linear operator}, it can be expressed as multiplication by a block-Toeplitz-with-Toeplitz-blocks (BTTB) matrix \cite{gray25_toeplitz}. 
This property means that the three sequential steps can be \emph{fused} into a single equivalent convolution step with an effective kernel matrix. 
Specifically, we rearrange the convolution kernel $W$ into a block operator.
Then, we flatten $W$ into a block matrix into $W \in \mathbb{R}^{r \times (r K^2)}$, and the resulting fused weight is: 
\begin{equation}
M_{\textsf{AWB}}' \;=\; AWB^{\!\top}, 
\end{equation}
where we can reshape $M_{\textsf{AWB}}' \in \mathbb{R}^{m \times (n K^2)}$ into $M_{\textsf{AWB}} \in \mathbb{R}^{m \times n \times K \times K}$, and the overall layer reduces to a standard convolution with kernel $M_{\textsf{AWB}} \in \mathbb{R}^{m \times (n K^2)}$: $X \;\xrightarrow{\;M_{\textsf{AWB}}\;} Y$. In practice, this means that instead of executing three steps (linear, convolution, and linear), we can pre-fuse the operations into one convolution, saving both computation and memory while preserving expressivity. 

\paragraph{Fuse then sine.}
\label{subsec:fusethensine}
As discussed in Sec.~\ref{subsec:sine}, the rank-enrichment property of the sine transformation only holds when it acts on a \emph{single low-rank matrix}. 
However, the sine function is not multiplicative homomorphic, i.e., $\mathrm{rank}\big(\sin(A W B)\big) \;\neq\; \mathrm{rank}\big(\sin(A)\sin(W)\sin(B)\big)$. In other words, applying sine transformation separately for each matrix doesn't preserve equivalence and effective rank expansion. Therefore, the transformation should be applied \emph{after} fusing $A$, $B$, and $W$ into $M_{\textsf{AWB}}$. 
Our \emph{task-specific} matrix is thus
\begin{equation}
M_t \;=\; \sin\!\big(\omega_t \cdot M_{\textsf{AWB}}\big) \;=\; \sin\!\big(\omega_t \cdot A W B^{\!\top}\big)
\label{eq:fuse_then_sine_simple}
\end{equation}
where $\omega_t$ is the task-specific frequency.
This \emph{fuse-then-sine} strategy guarantees that the nonlinearity acts directly on the shared base matrix, ensuring effective rank expansion. 

\begin{figure*}[t]
    \centering
    \begin{subfigure}[t]{0.3\textwidth}
        \centering
        \includegraphics[width=5.4cm]{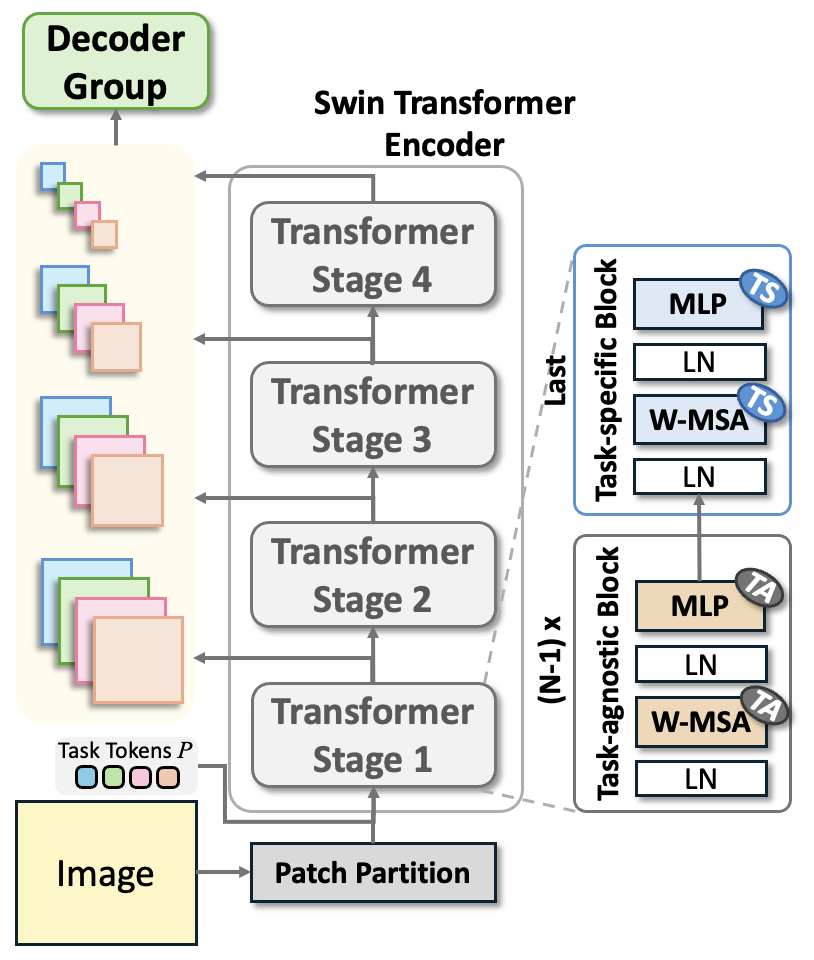}
        \caption{\textbf{Overall structure}}
        \label{fig:overall_structure}
    \end{subfigure}
    \hfill
    \begin{subfigure}[t]{0.18\textwidth}
        \centering
        \includegraphics[width=3.2cm]{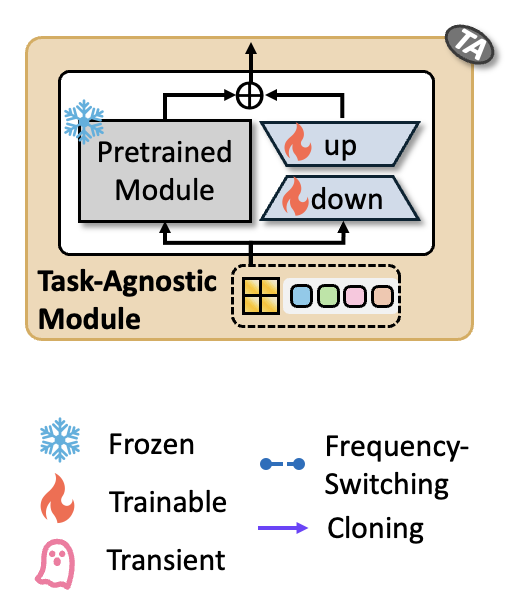}
        \caption{\textbf{Task-Agnostic Module}}
        \label{fig:ta}
    \end{subfigure}
    \begin{subfigure}[t]{0.48\textwidth}
        \centering
        \includegraphics[width=8cm]{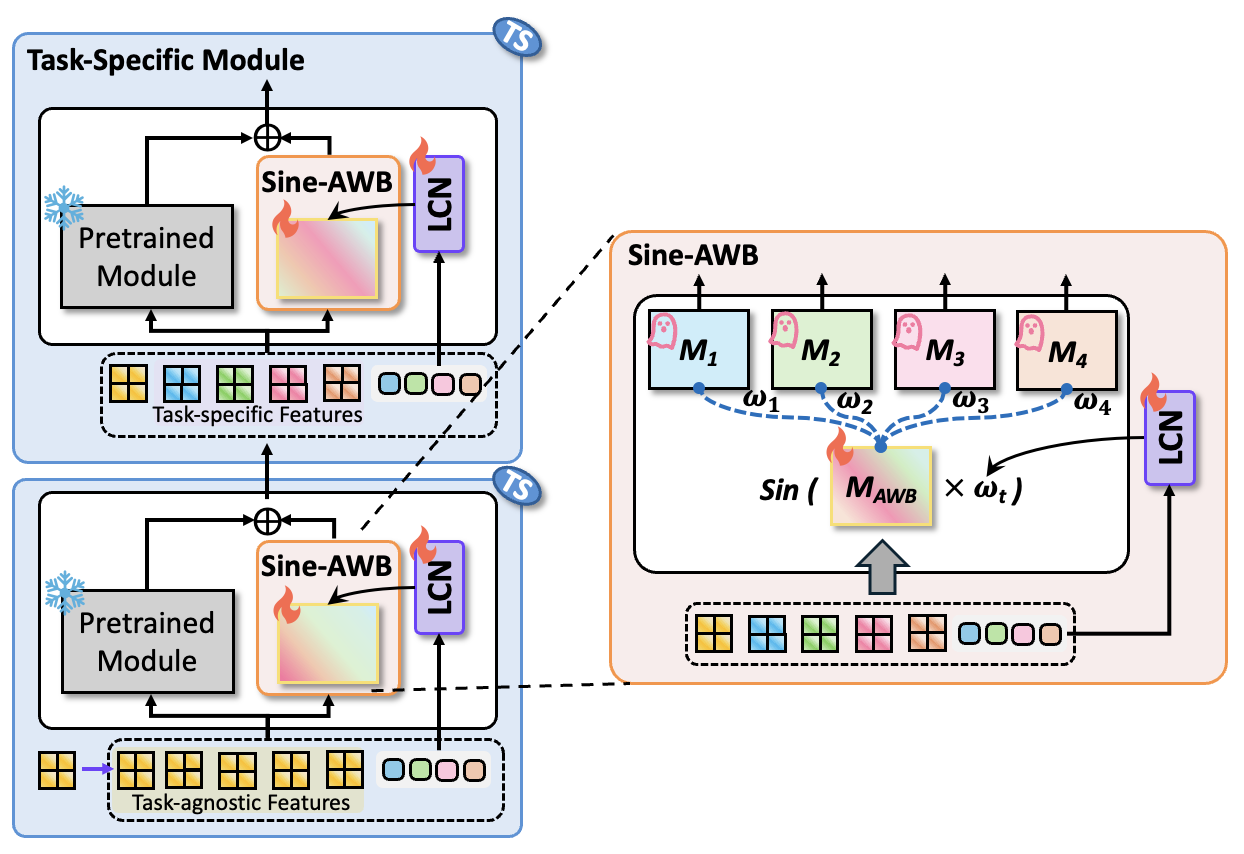}
        \caption{\textbf{Task-Specific Module}}
        \label{fig:ts}
    \end{subfigure}
    \caption{
    Overview of the Free Sinewich framework: 
    (a) A Swin Transformer Tiny \cite{liu2021Swin} serves as the shared encoder. The encoder receives image patch tokens with task tokens prepended to them. Following the VPT-shallow strategy \cite{jia2022vpt}, the task tokens are introduced only at the first Transformer stage.
    Task-agnostic features are extracted through the task-agnostic module (TA-Module) in all Transformer blocks except for the last block, while the last block employs a task-specific module (TS-Module) to capture task-dependent representations. 
    (b) Details of the TA-Module, which are basically the same as LoRA. 
    (c) The TS-Module includes a lightweight Clock Net (LCN) that takes task token $\boldsymbol{p}_t$ as input and determines a task frequency $\omega_t$. This frequency is used in the sine transformation to enhance the representational power of the low-rank adapter. Conceptually, we switch the shared base matrix $M_{\mathsf{AWB}}$ into different transient, task-specialized matrices $M_t$'s (ghost icon). 
    The transient $M_t$'s are instantiated on-the-fly to extract task-specific features, enabling efficient and scalable PEFT-MTL. 
    }
    \label{fig:overview}
\end{figure*}

\section{Methodology}

\subsection{Overview}

Figure~\ref{fig:overall_structure} shows the proposed Free Sinewich framework. Following TADFormer~\cite{baek2025tadformer}, a small bank of learnable \emph{task tokens} $P=\{\boldsymbol{p}_t\}_{t=1}^{T}$ is prepended to the image patch tokens $E=\{\boldsymbol{e}_k\}_{k=1}^{K}$ to form the input $X=[P,E]$ for a Swin Transformer Tiny~\cite{liu2021Swin} encoder. 
The encoder consists of four stages, each of which comprises $N$ blocks. The first $N-1$ blocks are task-agnostic blocks, and the last block is a task-specific block. 
In both types, a block comprises a layer normalization (LN), a window-based multi-head self-attention layer (W-MSA), followed by another LN layer and a multilayer perceptron (MLP). 

The task-agnostic blocks extract features generic to all tasks. Each layer in a task-agnostic block can be efficiently fine-tuned via the task-agnostic module (TA-module) shown in Figure~\ref{fig:ta}. A TA-module is a typical LoRA module. 

The task-specific block is the key to extracting features specific to different tasks. Each layer in this block is efficiently fine-tuned via the task-specific module (TS-module), where different frequencies are determined by a Clock Net depending on the task tokens. 
A frequency switching mechanism is developed to enhance the representational power of low-rank matrices. Details of the TS-module are shown in Figure~\ref{fig:ts}, and are elaborated in the following (Sec.~\ref{subsec:clock}, Sec.~\ref{subsec:sineawb_simple}, and Sec.~\ref{subsec:tsmodule}). 

\subsection{Lightweight Clock Net (LCN)}
\label{subsec:clock}
In the TS-Module, the LCN serves as a \emph{task-to-frequency translator} that assigns each task a distinct oscillation frequency ($\omega_t$),
driving different low-rank matrix transformations for different tasks.
Each task is represented by a learnable task token $\boldsymbol{p}_t \in \mathbb{R}^{C}$. 
A single-layer MLP $W_q:\mathbb{R}^{C}\!\to\!\mathbb{R}^{d_\omega}$ (shared across tasks) produces the task-dependent frequency:
\begin{align}
    \omega_t = s \cdot (\tanh(W_q \, \mathrm{ReLU}(\boldsymbol{p}_t)) + c),
    \label{eq:clocknet}
\end{align}
where $s$ and $c$ are learnable scale and offset parameters, respectively. 
The output $\omega_t$ of LCN is simply a scalar ($d_\omega = 1$), whose value is learned within an interval.
Conceptually, the task-dependent frequency determines how the shared base matrix \enquote{vibrates} into a task-specific form. 
In practice, LCN is not the main contributor to performance gains; it primarily generates bounded frequencies to stabilize training.

\subsection{Sine-AWB}
\label{subsec:sineawb_simple}
As shown in Figure~\ref{fig:ts}, the frequency $\omega_t$ is applied to enhance the shared base matrix $M_{\mathsf{AWB}}$. 
Following the fuse-then-sine process mentioned in Sec.~\ref{sec:prelim}, the enhanced matrix $M_t$ for the $t$-th task is obtained by 
\[
M_t \;=\; \sin\!\big(\omega_t \cdot M_{\mathsf{AWB}}\big). 
\]



With the adaptively-determined frequency and the sine transformation, a single shared base matrix $M_{\mathsf{AWB}}$ can be \emph{frequency-switched} into task-specific matrices $M_t$, $t=1, 2, ..., T$, where $T$ is the number of tasks to be solved simultaneously. 

Note that, although the TS-module (Figure~\ref{fig:ts}) is only illustrated for the encoder in Figure~\ref{fig:overall_structure}, it can also work for the decoder in exactly the same form. 

\paragraph{Low-pass filter.}
We observe that the sine-transformed matrix $M_t$ often comes with high-frequency noise. 
To smooth out these artifacts, we apply a $K{\times}K$ Gaussian low-pass filter to $M_t$. 
Formally, the filtered matrix $\widetilde{M}_t$ is obtained by:
\begin{equation}
    \widetilde{M}_t(x,y)
    = \sum_{u, v}
      G_\sigma(u,v)\, M_t(x{-}u, y{-}v),
    \label{eq:lowpass_conv}
\end{equation}
\begin{equation}
    G_\sigma(u,v)
    = \frac{1}{2\pi\sigma^2}
      \exp\!\left(-\frac{u^2 + v^2}{2\sigma^2}\right),
    \label{eq:gaussian_kernel}
\end{equation}
with standard deviation $\sigma{=}1$ and the kernel size $K{=}7$.  
This Gaussian filter is chosen for its simplicity and effectively suppresses high-frequency components while preserving structural details and improving stability (see Sec.~\ref{subsec:ablation}). 

\subsection{Frequency Switching Feature Processing}
\label{subsec:tsmodule}
Overall, the TS-Module applies the frequency determined by the LCN to the Sine-AWB module to produce task-specific features. 

Let $\boldsymbol{f}_i^t$ denote the intermediate feature map output by the $i$-th layer in the TS-module for the $t$-th task, and $\Phi_i(\cdot)$ the pre-trained transformation representing the process of the $i$-th layer. The transformation $\Phi_i(\cdot)$ is frozen when fine-tuning. The LCN takes the task token $\boldsymbol{p}_t$ and outputs a frequency value $\omega_t = \mathrm{LCN}(\boldsymbol{p}_t)$. Through the process described above, the smoothed matrix $\widetilde{M}_t$ acts as a convolutional kernel to perform channel-wise convolution with $\boldsymbol{f}_i^t$ to yield the task-specific output:
\begin{equation}
    \boldsymbol{f}_{i+1}^t = \Phi_i(\boldsymbol{f}_i^t) + (\widetilde{M}_t \ast \boldsymbol{f}_i^t),
    \label{eq:ts_forward}
\end{equation}
where $\ast$ denotes channel-wise convolution. This formulation is the same as LoRA, but the key is the enhanced low-rank matrix $\widetilde{M}_t$ obtained by the sine transformation. 

\subsection{Decoder Group}
\label{subsec:decoder_group}
Following prior PEFT-MTL frameworks~\cite{agiza2024mtlora,baek2025tadformer,mantri2025ditask}, for each task a set of multi-scale features $\{\boldsymbol{g}_1^t, \boldsymbol{g}_2^t, \boldsymbol{g}_3^t, \boldsymbol{g}_4^t\}$ are extracted by the shared encoder. The features from each stage are first projected by a task-specific $1{\times}1$ convolution.
The projected features of all stages are then upsampled to the same dimension $H \times W$, where $H$ and $W$ are the height and the width of the input image, respectively, and are concatenated as the fused representation $\boldsymbol{x}_t$: 
\begin{equation}
\tilde{\boldsymbol{g}}_i^t = 
\text{Up}_{H \times W}\big(\mathrm{Conv}_i^t(\boldsymbol{g}_i^t)\big),
\quad i = 1, \dots, 4,
\label{eq:decoder_fusion}
\end{equation}
\begin{equation}
\boldsymbol{x}_t = 
\mathrm{Concat}\big[
\tilde{\boldsymbol{g}}_1^t,\, 
\tilde{\boldsymbol{g}}_2^t,\, 
\tilde{\boldsymbol{g}}_3^t,\, 
\tilde{\boldsymbol{g}}_4^t
\big]. 
\label{eq:decoder_concat}
\end{equation}

Previous PEFT-MTL works~\cite{agiza2024mtlora,baek2025tadformer,mantri2025ditask} develop an independent decoder group $\Psi_{\mathrm{ind}} = \{\phi_1, \ldots, \phi_T\}$, where different decoders follow the same structure (e.g., HRNet~\cite{sun2019hrnet} or SegFormer~\cite{xie2021segformer}) but maintains an independent parameter set. 
Each decoder processes its corresponding task feature $\boldsymbol{x}_t$ in isolation, resulting in the total number of parameters scaling linearly with the number of tasks $T$ and introducing redundancy in both computation and memory usage. 

To address this issue, we propose a shared decoder group $\Psi_{\mathrm{shd}}$ that replaces $\Psi_{\mathrm{ind}}$ while keeping the same architecture. 
Taking HRNet as an example, each decoder $\phi_t$ in $\Psi_{\mathrm{ind}}$ follows a \texttt{Conv-BN-ReLU-Conv} structure, where the first convolutional layer alone contributes over one million parameters when replicated across $T$ tasks.
In contrast, $\Psi_{\mathrm{shd}}$ replaces these per-task convolution weights with the same set of shared weights $M_{\mathsf{AWB}}$, which is modulated into a task-specific form through the frequency-switching mechanism. 
\begin{equation}
\boldsymbol{h}_t = \widetilde{M}_t \ast \boldsymbol{x}_t + \boldsymbol{b}_t,
\label{eq:decoder_conv}
\end{equation}
where $\ast$ denotes channel-wise convolution, $\boldsymbol{b}_t$ is a task-specific bias vector.
The remaining decoder components include batch normalization, activation, and the final 
output convolution:
\begin{equation}
\hat{y}_t = \phi_t(\boldsymbol{h}_t),
\quad 
\phi_t = \texttt{BN-ReLU-Conv}.
\label{eq:decoder_output}
\end{equation}
This shared-decoder formulation retains the architectural integrity of the original 
decoder while transforming its main convolution into a frequency-switchable operator.  

\section{Experiments}

\begin{table*}[t]
\centering
\setlength{\tabcolsep}{6pt} 
\renewcommand{\arraystretch}{0.95} 
\small
\caption{\textbf{Comparison of parameter-efficient multi-task learning methods on the PASCAL-Context benchmark.} 
Using Swin Transformer Tiny pretrained on the ImageNet-1K dataset as the backbone,
$\Delta m$ denotes the average relative improvement over the single-task baseline (positive indicates better), and $\uparrow$ / $\downarrow$ indicate higher / lower is better. 
Across all ranks, Free Sinewich delivers the best trade-off between performance and efficiency.
* means the method uses the weights of Swin Transformer Tiny pretrained on the ImageNet-22k~\cite{imagenet} dataset.
}
\begin{tabular}{c|cccc|c|c}
\toprule
\multirow{2}{*}{\textbf{Method}} & \textbf{SemSeg} & \textbf{Human Parts} & \textbf{Saliency} & \textbf{Normals} & \multirow{2}{*}{\textbf{$\Delta m$ (\%)}} & \textbf{Trainable}  \\
 & (mIoU ↑) & (mIoU ↑) & (mIoU ↑) & (rmse ↓) & & \textbf{Parameters (M)}  \\
\midrule
Single Task & 67.21 & 61.93 & 62.35 & 17.97 & 0 & 112.62  \\
MTL - Tuning Decoders Only & 65.09 & 53.48 & 57.46 & 20.69 & -9.95 & 1.94  \\
MTL - Full Fine Tuning & 67.56 & 60.24 & 65.21 & 16.64 & +2.23 & 30.06  \\
\midrule
Adapter~\cite{he2022adapter} & 69.21 & 57.38 & 61.28 & 18.83 & -2.71 & 11.24  \\
Bitfit~\cite{ben2022bitfit} & 68.57 & 55.99 & 60.64 & 19.42 & -4.60 & 2.85  \\
VPT-shallow~\cite{jia2022vpt} & 62.96 & 52.27 & 58.31 & 20.90 & -11.18 & 2.57  \\
VPT-deep~\cite{jia2022vpt} & 64.35 & 52.54 & 58.15 & 21.07 & -10.85 & 3.43  \\
Compactor~\cite{compacter} & 68.08 & 56.41 & 60.08 & 19.22 & -4.55 & 2.78  \\
Compactor++~\cite{compacter} & 67.26 & 55.69 & 59.47 & 19.54 & -5.84 & 2.66  \\
LoRA~\cite{hu2022lora} & 70.12 & 57.73 & 61.90 & 18.96 & -2.17 & 2.87  \\
VL-Adapter~\cite{sung2022vladapter} & 70.21 & 59.15 & 62.29 & 19.26 & -1.83 & 4.74  \\
HyperFormer~\cite{karimi2021hyperformer} & 71.43 & 60.73 & 65.54 & 17.77 & +2.64 & 72.77  \\
Polyhistor~\cite{liu2022polyhistor} & 70.87 & 59.54 & 65.47 & 17.47 & +2.34 & 8.96  \\
MTLoRA~\cite{agiza2024mtlora} ($r=16$) & 68.19 & 58.99 & 64.48 & 17.03 & +1.35 & 4.95  \\
MTLoRA~\cite{agiza2024mtlora} ($r=32$) & 67.74 & 59.46 & 64.90 & 16.59 & +2.16 & 6.08  \\
MTLoRA~\cite{agiza2024mtlora} ($r=64$) & 67.90 & 59.84 & 65.40 & 16.60 & +2.55 & 8.34  \\
DiTASK~\cite{mantri2025ditask} - MTL & 70.09 & 59.03 & 64.55 & 17.47 & +1.47 & 3.55  \\
DiTASK~\cite{mantri2025ditask} - MTL* & 69.66 & 62.02 & 65.00 & 17.10 & +3.22 & 3.55  \\
TADFormer~\cite{baek2025tadformer} ($r=16$) & 69.79 & 59.27 & 65.04 & 16.91 & +2.44 & 3.56  \\
TADFormer~\cite{baek2025tadformer} ($r=32$) & 70.20 & 60.00 & 65.71 & 16.57 & +3.63 & 4.78  \\
TADFormer~\cite{baek2025tadformer} ($r=64$) & 70.82 & 60.45 & 65.88 & 16.48 & +4.24 & 7.38  \\
\midrule
\textbf{Free Sinewich ($r=16$)} & 70.92 & 59.78 & 65.32 & 16.70 & \textbf{+3.47} & \textbf{2.79}  \\
\textbf{Free Sinewich ($r=32$)} & 71.02 & 60.75 & 65.94 & 16.44 & \textbf{+4.51} & \textbf{4.04}  \\
\textbf{Free Sinewich ($r=64$)} & 71.25 & 61.38 & 66.24 & 16.14 & \textbf{+5.39} & \textbf{6.53}  \\
\bottomrule
\end{tabular}
\label{tab:main_mtl}
\end{table*}

\begin{table*}[t]
\centering
\caption{\textbf{Performance comparison on the NYUDv2~\cite{nyudv2} dataset.}
All methods utilize the Swin Transformer Tiny~\cite{liu2021Swin} backbone pre-trained on the ImageNet-22K~\cite{imagenet} dataset, with HRNet as the decoder for fair comparison. }
\setlength{\tabcolsep}{6pt} 
\renewcommand{\arraystretch}{0.95} 
\small
\begin{tabular}{l|cccc|c|c}
\toprule
\multirow{2}{*}{\textbf{Method}} & \textbf{SemSeg} & \textbf{Depth} & \textbf{Edge} & \textbf{Normals} & \multirow{2}{*}{\textbf{$\Delta m$ (\%)}} & {\textbf{Trainable}} \\
 & (mIoU ↑) & (rmse ↓) & (odsF ↑) & (rmse ↓) &  & \textbf{Parameters (M)} \\
\midrule
Single Task & 42.59 & 66.08 & 59.80 & 22.58 & 0 & 112.64 \\
MTL - Tuning Decoders Only & 34.34 & 84.47 & 57.40 & 32.93 & -24.26 & 1.96 \\
MTL - Full Fine Tuning & 42.07 & 65.36 & 58.80 & 23.74 & -1.73 & 30.08 \\
\midrule
MTLoRA~\cite{agiza2024mtlora} ($r=64$) & 41.11 & 65.49 & 58.30 & 24.55 & -3.45 & 8.36 \\
DiTASK~\cite{mantri2025ditask} - MTL & 37.36 & 75.00 & 57.10 & 28.63 & -14.27 & 3.57 \\
TADFormer~\cite{baek2025tadformer} ($r=64$) & 41.37 & 64.10 & 58.50 & 24.54 & -2.68 & 7.40 \\
\midrule
Free Sinewich ($r=16$) & 40.90 & 67.15 & 58.50 & 25.06 & -4.68 & 2.81 \\
Free Sinewich ($r=32$) & 41.80 & 66.67 & 58.80 & 24.67 & -3.41 & 4.06 \\
Free Sinewich ($r=64$) & 42.26 & 64.08 & 59.40 & 23.41 & \textbf{-0.52} & \textbf{6.55} \\
\bottomrule
\end{tabular}
\label{tab:nyud_results}
\end{table*}

\begin{table*}[t]
\centering
\caption{\textbf{Performance variations when different components are used.}
Removing key components such as the LCN, the low-pass filter, or the sine transformation leads to performance degradation, confirming that each element contributes to overall performance and efficiency. }
\setlength{\tabcolsep}{8pt} 
\renewcommand{\arraystretch}{0.95} 
\small
\begin{tabular}{l|cccc|c|c}
\toprule
\multirow{2}{*}{\textbf{Method}} & \textbf{SemSeg} & \textbf{Human Parts} & \textbf{Saliency} & \textbf{Normals} & \multirow{2}{*}{\textbf{$\Delta m$ (\%)}} & {\textbf{Trainable}} \\
 & (mIoU ↑) & (mIoU ↑) & (mIoU ↑) & (rmse ↓) &  & \textbf{Parameters (M)} \\
\midrule
\textbf{Free Sinewich (Ours)} & \textbf{71.25} & \textbf{61.38} & \textbf{66.24} & \textbf{16.14} & \textbf{+5.39} & \textbf{6.53} \\
\midrule
Ours w/o LCN & 70.83 & 61.37 & 66.09 & 16.17 & +5.12 & 6.51 \\
Ours w/o Low-pass filter & 70.95 & 61.33 & 65.44 & 16.22 & +4.82 & 6.53 \\
Ours w/o Sine & 69.68 & 60.69 & 64.91 & 16.37 & +3.67 & 6.53 \\
\bottomrule
\end{tabular}
\label{tab:ablation}
\end{table*}

\begin{table*}[t]
\centering
\caption{\textbf{Performance variations when different configurations are used for the decoder.}
All methods use the Swin Transformer Tiny backbone pre-trained on the ImageNet-22K dataset, with rank $r{=}32$ for a fair comparison on PASCAL-Context dataset.
}
\small
\setlength{\tabcolsep}{4pt}
\begin{tabular}{l l | c c c c | c | c}
\toprule
\multirow{2}{*}{\textbf{Method}} &\multirow{2}{*}{\textbf{Decoder}}& \textbf{SemSeg} & \textbf{Human Parts} & \textbf{Saliency} & \textbf{Normals} & \multirow{2}{*}{\textbf{$\Delta m$ (\%)}} & {\textbf{Trainable Param.} (M)} \\
& & (mIoU ↑) & (mIoU ↑) & (mIoU ↑) & (rmse ↓) &  & Decoder / All \\
\midrule
\multirow{3}{*}{TADFormer~\cite{baek2025tadformer}} & HRNet~\cite{sun2019hrnet}  & 72.05 & 61.60 & 65.45 & 16.70 & +4.67 & 1.94 / 4.78 \\
          & SegFormer~\cite{xie2021segformer}  & 72.33 & 61.16 & 65.80 & 16.87 & +4.51 & 2.08 / 4.91 \\
          & ASPP~\cite{aspp}  & 73.66 & 60.37 & 65.27 & 16.43 & +5.09 & 12.44 / 15.27 \\
\midrule
\multirow{3}{*}{Free Sinewich} & HRNet~\cite{sun2019hrnet} & 72.65 & 62.57 & 65.55 & 16.36 & +5.80 & 1.07 / 4.04 \\
          & SegFormer~\cite{xie2021segformer}  & 72.68 & 62.67 & 66.16 & 16.53 & +5.86 & 1.08 / 4.05 \\
          & ASPP~\cite{aspp}  & 75.91 & 61.92 & 65.84 & 16.17 & +7.14 & 6.64 / 9.61 \\
\bottomrule
\end{tabular}
\label{tab:decoder_comparison}
\end{table*}

\begin{table*}[t]
\centering
\caption{
\textbf{Performance variations when different sharing schemes are employed.}
Performance comparison among the shared base (Free Sinewich), independent base, and independent decoder variants on the PASCAL-Context dataset. Sharing a single base matrix across tasks consistently improves the overall performance with fewer parameters. 
}
\small
\setlength{\tabcolsep}{4pt}
\begin{tabular}{l|cccc|c|c}
\toprule
\multirow{2}{*}{\textbf{Method}} & \textbf{SemSeg} & \textbf{Human Parts} & \textbf{Saliency} & \textbf{Normals} & \multirow{2}{*}{\textbf{$\Delta m$ (\%)}} & {\textbf{Trainable}} \\
 & (mIoU ↑) & (mIoU ↑) & (mIoU ↑) & (rmse ↓) &  & \textbf{Parameters (M)} \\
\midrule
\textbf{Shared Base} & \textbf{71.25} & \textbf{61.38} & \textbf{66.24} & \textbf{16.14} & \textbf{+5.39} & \textbf{6.53} \\
\midrule
Independent Base  & 70.81 & 61.56 & 65.42 & 16.09 & +5.03 & 10.22 \\
Independent Decoder & 70.91 & 61.57 & 66.03 & 16.10 & +5.31 & 7.41 \\
\bottomrule
\end{tabular}
\label{tab:sine_ablation}
\end{table*}

\subsection{Experimental Settings}

\label{subsec:data}
\paragraph{Datasets and Tasks.}
We evaluate our method on two standard multi-task benchmarks: \textbf{PASCAL-Context}~\cite{everingham2010pascal} and \textbf{NYUDv2}~\cite{nyudv2}.
PASCAL-Context contains 4{,}998 training images and 5{,}105 validation images with annotations for semantic segmentation (21 classes), human part segmentation (7 classes), saliency detection, and surface normal estimation.
NYUDv2 consists of 1{,}449 RGB-D images from 464 indoor scenes, split into 795 training samples and 654 testing samples.
It supports four dense prediction tasks: semantic segmentation (40 classes), monocular depth estimation, surface normal estimation, and edge detection.

\paragraph{Evaluation Metrics.}
Following the setting of \cite{vandenhende2020mti}, we report mIoU for semantic segmentation, human parts, and saliency; and angular root mean squared error (RMSE) in degrees for surface normals. 
On NYUDv2~\cite{nyudv2}, we additionally report depth RMSE and edge ODSF (Optimal Dataset Scale F-measure). 
To summarize overall effectiveness across tasks we use $\Delta m$, the average relative improvement over a single-task baseline:
\begin{equation}
\label{eq:delta_m}
\Delta m \;=\; \frac{1}{T}\sum_{t=1}^{T} (-1)^{\ell_t} \frac{R^{\text{MTL}}_t - R^{\text{ST}}_t}{R^{\text{ST}}_t},
\end{equation}
where $T$ is the number of tasks, $R^{\text{ST}}_t$ and $R^{\text{MTL}}_t$ denote single-task and multi-task performance of the $t$-th task, respectively. The value $\ell_t{=}1$ if a lower value means better performance (e.g., RMSE), otherwise $0$.


\paragraph{Training.}
Each input image is associated with $T$ task-specific ground-truth outputs. 
Following the standard multi-task training protocol in the prior work~\cite{vandenhende2020mti}, all tasks are jointly optimized during training. 
For each mini-batch, predictions for all $T$ tasks are computed, and their losses are aggregated to form the overall objective:
\begin{equation}
\mathcal{L}_{\text{MTL}} = \sum_{t=1}^{T} w_t\,\mathcal{L}_t,
\end{equation}
where $\mathcal{L}_t$ and $w_t$ denote the loss and weight of the $t$-th task, respectively. 
Gradients of the loss are back-propagated through the shared encoder and task-specific decoders, and parameters are updated once per batch. 
Task weights and loss terms are set as in~\cite{vandenhende2020mti}.


\subsection{Baselines}
\label{subsec:baselines}
We compare performance and trainable parameters with several PEFT-MTL baselines: 
\textbf{Single Task} denotes a baseline where each task is trained by fully fine-tuning its own independently pretrained model. 
\textbf{MTL-Tuning Decoders Only} is a method where the shared encoder is frozen, and only the task-specific decoders are trainable. 
\textbf{MTL-Full Fine-Tuning} is a method where both the shared encoder and the task-specific decoder are trainable.
\textbf{Adapter}~\cite{he2022adapter} and \textbf{BitFit}~\cite{ben2022bitfit} modify small subsets of parameters. 
\textbf{VPT}~\cite{jia2022vpt} tunes learnable prompts. 
\textbf{Compactor}~\cite{compacter} uses low-rank/Kronecker reparameterizations. 
\textbf{LoRA}~\cite{hu2022lora} applies low-rank updates to attention and MLP. 
\textbf{VL-Adapter}~\cite{sung2022vladapter} shares a single adapter across tasks. 
\textbf{HyperFormer}~\cite{karimi2021hyperformer} and \textbf{Polyhistor}~\cite{liu2022polyhistor} generate adapters via hypernetworks. 
\textbf{MTLoRA}~\cite{agiza2024mtlora} splits updates into task-agnostic and task-specific LoRA branches. 
\textbf{TADFormer}~\cite{baek2025tadformer} injects task-aware prompts with dynamic task filters. 
\textbf{DiTASK}~\cite{mantri2025ditask} performs differentiable spectral transforms on singular values to adapt tasks.

\subsection{Quantitative Analysis}
\label{subsec:overall}

\paragraph{Main Results.}
Table~\ref{tab:main_mtl} reports the performance along with the corresponding numbers of trainable parameters. 
For a fair comparison, all models utilize the Swin Transformer Tiny backbone~\cite{liu2021Swin}, pretrained on the ImageNet-1K dataset. 
We see that our proposed Free Sinewich gives better performance while maintaining a lower number of trainable parameters. 
Comparing with the latest baselines, when the rank of the weight matrix is the same, Free Sinewich ($r{=}64$) achieves mIoUs of 71.25 (SemSeg), 61.38 (Human Parts), 66.24 (Saliency), and RMSE of 16.14 (Normal prediction), achieving average +5.39 $\Delta m$ with only 6.53M trainable parameters. These performances surpass previous state-of-the-art TADFormer~\cite{baek2025tadformer} ($r{=}64$), which achieves +4.24 $\Delta m$ with more parameters (7.38M). 
Notably, even with a lower rank ($r{=}32$), Free Sinewich achieves performance better than TADFormer ($r{=}64$), indicating the effectiveness of the frequency-switching sine modulation. 


\paragraph{NYUDv2 Results.}
Table~\ref{tab:nyud_results} presents the performance on the NYUDv2 dataset. 
Free Sinewich consistently achieves strong results across all four tasks while maintaining a compact parameter budget. 
With rank $r{=}64$, Free Sinewich attains a near-zero $\Delta m$ of -0.52, effectively matching the full fine-tuning setting with only 6.55M trainable parameters. Even with smaller ranks ($r{=}16$ and $r{=}32$), the performance remains competitive, demonstrating that the frequency-switching modulation captures task-specific variations efficiently. 
Compared to prior PEFT-MTL approaches, Free Sinewich achieves a superior balance between performance and parameter efficiency, highlighting its effectiveness for compact multi-task learning on the NYUDv2 benchmark. 

\subsection{Ablation Studies}
\label{subsec:ablation}
\paragraph{Influence of Different Components.}
Table~\ref{tab:ablation} evaluates the contribution of LCN, the low-pass filter, and the sine transformation. 
Removing the sine transformation leads to the largest degradation (from +5.39 to +3.67 $\Delta m$), confirming that the frequency switching mechanism on a shared base matrix is the primary driver of performance gains. The LCN and the low-pass filter further stabilize training and enhance the quality of semantic segmentation and saliency estimation. 
\label{subsec:decoders}

\paragraph{Decoder Robustness.}
We pair TADFormer~\cite{baek2025tadformer}, and Free Sinewich with three different decoders, including HRNet~\cite{sun2019hrnet}, SegFormer~\cite{xie2021segformer}, and ASPP~\cite{aspp}. 
Table~\ref{tab:decoder_comparison} shows performance variations when different decoders are used. As can be seen, Free Sinewich consistently yields higher $\Delta m$'s no matter which decoder is used. With ASPP~\cite{aspp} as the decoder, it achieves the best $\Delta m$ (+7.14). 


\paragraph{Ablation on Parameter Sharing.}
To test whether frequency switching truly enables parameter reuse across tasks, we compare three variants that remove cross-task sharing in Sine-AWB.  
The first variant is the \textbf{Shared-Base} setting (corresponding to our Free Sinewich). All tasks use a single shared base matrix $M_{\mathsf{AWB}}$, meaning that $M_t = \sin(\omega_t \cdot M_{\mathsf{AWB}})$ modulates the same parameters with different frequencies. 
The second variant is the \textbf{Independent-Base} setting, which gives each task its own LoRA factors $A^t$, $B^t$, and kernel $W^t$, forming $M^t_{\mathsf{AWB}} = A^t W^t {B^t}^{\top}$ with no parameter shared across tasks.  
The third variant is the \textbf{Independent-Decoder} setting, which implements independent decoder groups $\Psi_{\mathrm{ind}} = \{\phi_1, \ldots, \phi_T\}$, where each $\phi_t$ follows the HRNet~\cite{sun2019hrnet} structure but maintains its own parameters (see Sec.~\ref{subsec:decoder_group}). In this setting, each task employs its own decoder, removing cross-task parameter sharing. 
Table~\ref{tab:sine_ablation} shows that, without sharing the base matrix, the performance drops, and more parameters are needed. 
This confirms that performance gains stem from the shared parameter and the frequency-based specialization. 


\begin{figure}[t]
    \centering
    \includegraphics[width=0.9\linewidth]{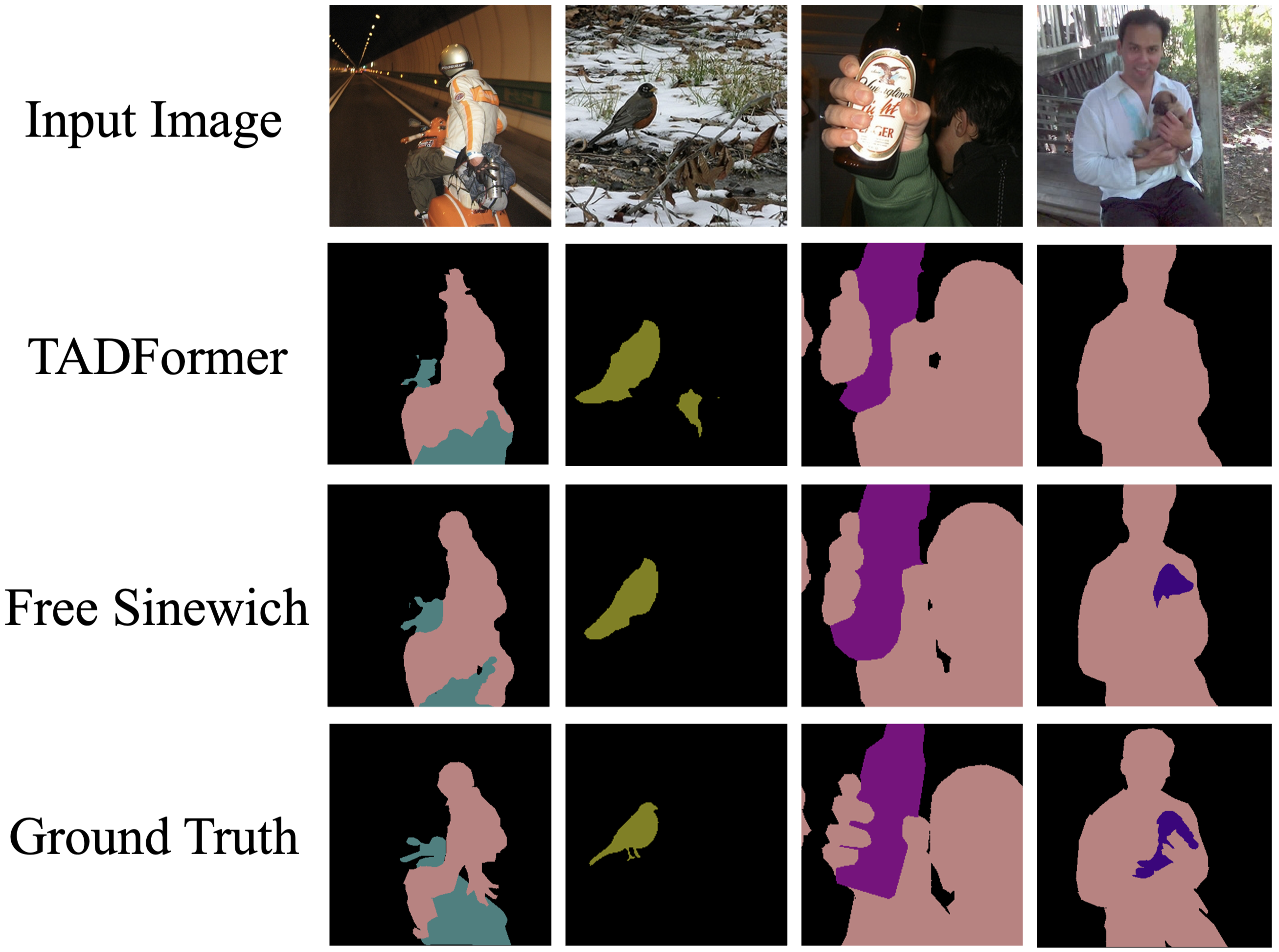}
    \caption{
    \textbf{Visual comparison of semantic segmentation on the Pascal-Context dataset.} The Free Sinewich method yields fuller and more coherent segmentation masks, capturing finer-grained boundary details than TADFormer.} 
    \label{fig:qualitative}
\end{figure}

\subsection{Qualitative Analysis}
\label{subsec:qualitative}

Figure~\ref{fig:qualitative} shows a qualitative comparison of semantic segmentation on the validation set. 
Compared to TADFormer at the same rank, Free Sinewich preserves sharper semantic boundaries with fewer label leaks, achieves higher task precision with reduced background activation, and generates smoother segmentation maps. 
These trends align with the quantitative gains in Tables~\ref{tab:main_mtl}.

\section{Related Works}

\subsection{Parameter-Efficient Fine-Tuning}
PEFT methods aim to adapt large pre-trained models with minimal additional parameters while approximating the representational capacity. Early techniques such as Adapters~\cite{he2022adapter}, BitFit~\cite{ben2022bitfit}, and IA$^3$~\cite{liu2022ia3} introduced lightweight residual modules or bias tuning to efficiently steer frozen backbones. 
Low-rank adaptation became a dominant approach through LoRA~\cite{hu2022lora} and its variants (AdaLoRA~\cite{zhang2023adalora}, QLoRA~\cite{dettmers2023qlora}), which decompose the weight matrix into lower-rank matrix multiplication. 
Subsequent research further enhanced LoRA's expressivity: DoRA~\cite{liu2024dora} decouples direction and magnitude to improve rank flexibility, HiRA~\cite{huang2025hira} hierarchically composes low-rank layers for higher effective rank, and ConvLoRA~\cite{zhong2024convlora} integrates convolutional priors into low-rank modules for vision tasks. 
While these PEFT methods enhance the \emph{expressive power} of low-rank adapters, they were primarily designed for single-task adaptation. In contrast, our approach explores low-rank adapters for multiple tasks simultaneously. 

\subsection{PEFT for MTL}
To enhance efficiency in MTL, several studies have adopted the PEFT paradigm for MTL. 
MTLoRA~\cite{agiza2024mtlora} extends LoRA by decomposing low-rank adapters into task-agnostic and task-specific components, balancing knowledge sharing and task specialization. 
DiTask~\cite{mantri2025ditask} introduces differentiable homomorphic transformations on singular values to adapt pretrained weights for each task while preserving the shared subspace. 
TADFormer~\cite{baek2025tadformer} incorporates task-aware prompts and dynamic task filters to inject image priors for dense prediction without retraining the entire backbone. 
Similarly, Polyhistor~\cite{liu2022polyhistor} leverages layer-wise decomposed hypernetworks for adaptive task modulation. 
While these PEFT-MTL frameworks achieve impressive efficiency and flexibility, they are limited to \emph{pseudo-share} low-rank adapters through multiple paths rather than genuine parameter reuse within the shared backbone. 
This motivates our proposed approach, which adopts frequency-based modulation to enable the true sharing of low-rank adapters across multiple tasks.

\section{Conclusion}

We have presented Free Sinewich, a parameter-efficient framework for multi-task learning that achieves \textit{true weight reuse} by the frequency switching mechanism. 
To extract task-specific features, we employ a lightweight Clock Net to determine task-dependent frequencies and a Sine-AWB layer that applies an element-wise sine transformation based on the determined frequencies. The sine transformation increases the effective rank of low-rank adaptors, and the frequency switching mechanism decorrelates weights, mitigating interference between different tasks.
Extensive experiments on two benchmarks demonstrate that the proposed methods yield strong performance-efficiency trade-offs.
Ablation studies also confirm that both the Clock Net and the sine modulation are necessary for the observed gains. 
Future work includes learning spatially/temporally varying frequencies, extending to video and multimodal settings, and integrating frequency conditioning with dynamic routing.

\noindent\textbf{Acknowledgement.}
This work was funded in part by the National Science and Technology Council, Taiwan, under grants 114-2622-E-006-028, 114-2221-E-006-047-MY3, 112-2221-E-006-136-MY3, 115-2425-H-006-005, and 114-2634-F-006-002.
{
    \small
    \bibliographystyle{ieeenat_fullname}
    \bibliography{main}
}
\clearpage
\twocolumn[
\begin{center}
\Large\textbf{Supplementary Material}
\end{center}
\vspace{0.5em}
]
\setcounter{section}{0}
\renewcommand{\thesection}{\Alph{section}}
\section{Theoretical Analysis of Task Decorrelation}

In this section, we provide a theoretical analysis showing that the proposed
\emph{frequency-switching} sine modulation decorrelates task-specific matrices derived
from a shared low-rank base, whereas linear frequency scaling alone cannot.
This analysis formally explains why the sine transformation is a necessary
component for effective task specialization in our framework.

\paragraph{Correlation Definition.}
For two matrices $M_s, M_t \in \mathbb{R}^{m\times n}$, we define their
correlation as the cosine similarity of their vectorized forms:
\begin{equation}
\mathrm{corr}(M_s,M_t)
=
\frac{
    \mathrm{vec}(M_s)^\top \mathrm{vec}(M_t)
}{
    \|\mathrm{vec}(M_s)\|_2 \;
    \|\mathrm{vec}(M_t)\|_2
}.
\label{eq:corr}
\end{equation}
A value of $0$ indicates decorrelation (orthogonality), while
$\pm1$ indicates perfectly correlated matrices.

\paragraph{Proposition 1.} 
\textit{
Let $M_{\mathrm{AWB}} = A W B^\top$ be the shared base matrix.
Define task-specific kernels using frequency-dependent sine modulation:
\begin{equation}
M_t = \sin(\omega_t M_{\mathrm{AWB}}), 
\qquad t = 1,\dots,T,
\end{equation}
where the sine is applied elementwise.
Assume each entry of $M_{\mathrm{AWB}}$ is a zero-mean,
finite-variance random variable with symmetric density.
Then for any two distinct frequencies $\omega_s \neq \omega_t$,
we have}
\begin{equation}
\mathrm{corr}(M_s, M_t) \approx 0,
\end{equation}
\textit{i.e., frequency-dependent sine mapping decorrelates the resulting
task-specific matrices.}
  
\paragraph{Proof.}
Let $X$ be a scalar entry of $M_{\mathrm{AWB}}$, and define
$Y_s=\sin(\omega_s X)$ and $Y_t=\sin(\omega_t X)$.
Using the identity
$
\sin(\alpha X)\sin(\beta X)
=\tfrac{1}{2}\big(\cos((\alpha-\beta)X)-\cos((\alpha+\beta)X)\big),
$
we obtain
\[
\mathbb{E}[Y_s Y_t]
= \tfrac{1}{2}\big(
\mathbb{E}[\cos((\omega_s-\omega_t)X)]
-
\mathbb{E}[\cos((\omega_s+\omega_t)X)]
\big).
\]

Because \(X\) has a symmetric density and finite variance, the expectation
\(\mathbb{E}[\cos(kX)]\) goes to zero as \(|k|\to\infty\).
Therefore, whenever both $|\omega_s-\omega_t|$ and $|\omega_s+\omega_t|$
are large, the two cosine expectations are small, and thus
\[
\mathbb{E}[Y_s Y_t] \approx 0.
\]

Vectorizing $M_t$ gives $\mathbf{m}_t=\mathrm{vec}(M_t)$, whose entries
satisfy $M_t(i,j)\sim Y_t$, where $Y_t=\sin(\omega_t X)\in[-1,1]$.
By the law of large numbers,
\begin{equation}
\frac{\mathbf{m}_s^\top\mathbf{m}_t}{\|\mathbf{m}_s\|_2\|\mathbf{m}_t\|_2}
\;\xrightarrow[]{}
\;
\frac{\mathbb{E}[Y_sY_t]}{\sqrt{\mathbb{E}[Y_s^2]}\sqrt{\mathbb{E}[Y_t^2]}}
\;\approx\;0,
\end{equation}
which proves $\mathrm{corr}(M_s,M_t)\approx 0$.
\hfill $\square$
\paragraph{Proposition 2.} 
\textit{
Without sine transformation, frequency acts only as a scalar scaling and thus cannot decorrelate the AWB matrix.
Define the linear-scaled task matrices}
\begin{equation}
\widetilde{M}_t = \omega_t M_{\mathrm{AWB}}.
\end{equation}
\textit{
Then for any tasks $s,t$ we have}
\begin{equation}
|\mathrm{corr}(\widetilde{M}_s,\widetilde{M}_t)| = 1,
\end{equation}
\textit{
i.e., the task-specific matrices are perfectly correlated and lie in the
same one-dimensional subspace.}

\paragraph{Proof.}
Vectorizing yields
\begin{equation}
\widetilde{\mathbf{m}}_t
=
\mathrm{vec}(\widetilde{M}_t)
=\omega_t\,\mathrm{vec}(M_{\mathrm{AWB}})
=\omega_t \mathbf{m}.
\end{equation}
For any $s,t$,
\begin{equation}
\mathrm{corr}(\widetilde{M}_s,\widetilde{M}_t)
=
\frac{
(\omega_s\mathbf{m})^\top(\omega_t\mathbf{m})
}{
|\omega_s|\|\mathbf{m}\|_2 \; |\omega_t|\|\mathbf{m}\|_2
}
\textcolor{blue}{}=\pm 1.
\end{equation}
Thus all $\widetilde{M}_t$ are collinear scalings of the same matrix and share identical singular directions. 
Hence frequency without sine cannot create decorrelated or task-specific variations.
\hfill $\square$

\paragraph{Implications.}
Together, Propositions~1 and~2 formally establish that the sine-based frequency modulation is the key mechanism enabling task-specific decorrelation from a shared low-rank base. 
Frequency alone, without the sine transformation, collapses all task-specific matrices into the same parameter subspace. 
This theoretical result directly supports the design of our frequency-switching Sine-AWB module and explains the empirical performance gains observed in our experiments.

\begin{figure}[t]
    \centering
    \begin{minipage}{0.9\linewidth}
    \subfloat[Effect of pretraining dataset.]{
        \includegraphics[width=0.95\linewidth]{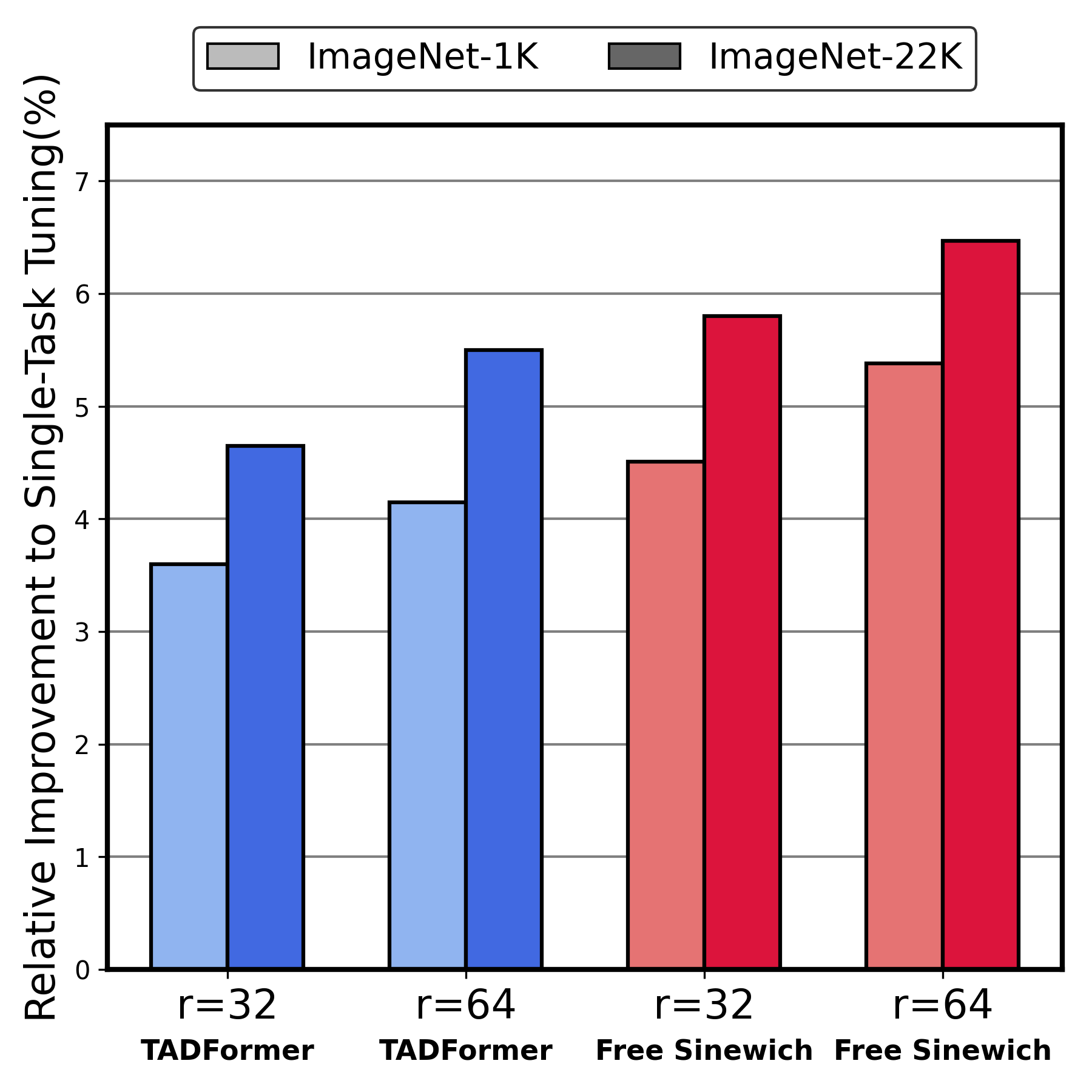}
        \label{fig:pretraining}
    }
    \hfill
    \subfloat[Effect of backbone capacity.]{
        \includegraphics[width=0.95\linewidth]{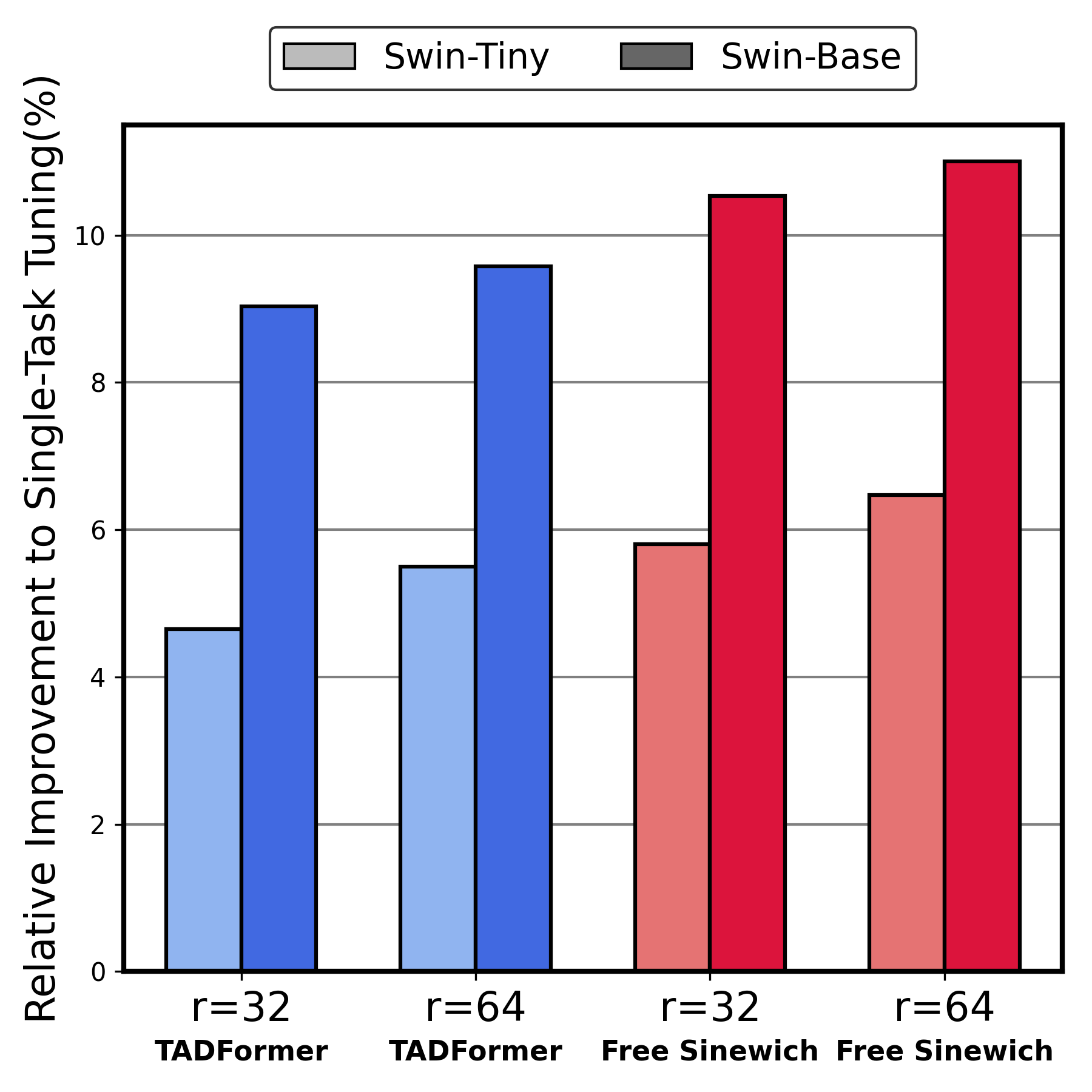}
        \label{fig:backbone_size}
    }
    \caption{\textbf{Effect of backbone pretraining and model capacity on Free Sinewich.}
    (a) Comparison between Swin-T pretrained on ImageNet-1K and ImageNet-22K.
    (b) Performance comparison between Swin-T and Swin-B backbones (both pretrained on ImageNet-22K).}
    \label{fig:backbone_pretraining_and_size}
    \end{minipage}
\end{figure}

\section{Free Sinewich with Different Backbones and Pretraining Datasets}

We evaluate the robustness of Free Sinewich under different backbone capacities and pretraining settings.
Specifically, we examine the effect of (i) using a larger pretraining dataset and (ii) adopting a larger backbone architecture.
The results are summarized in Fig.~\ref{fig:backbone_pretraining_and_size}.

As shown in Fig.~\ref{fig:pretraining}, pretraining on a larger dataset (ImageNet-22K)
leads to a clear and consistent performance improvement over ImageNet-1K pretraining.
This indicates that a stronger pretrained backbone provides a richer shared representation,
which directly enhances the effectiveness of frequency switching.
Since Free Sinewich reuses a single shared base matrix across tasks,
the quality of the pretrained weights plays a crucial role in enabling effective task-specific modulation.

Fig.~\ref{fig:backbone_size} further shows that increasing backbone capacity
from Swin-T to Swin-B also improves performance.
This confirms that Free Sinewich benefits from stronger representational capacity in the shared backbone.
The relative gain from enlarging the backbone is bigger for Free Sinewich than for TADFormer.
We attribute this to a effect:
Free Sinewich already achieves strong performance by efficiently reusing shared parameters through frequency switching,
this further scales up with a bigger model.
These results demonstrate that Free Sinewich scales favorably with both better pretraining and larger backbones,
while maintaining its core advantage of parameter-efficient task specialization through frequency-based modulation.

\section{Gaussian Low-Pass Filter Hyperparameters}
\begin{table*}[t]
\centering
\caption{\textbf{Ablation over low-pass filter hyperparameters.}
Effect of kernel size $K$ and standard deviation $\sigma$ on performance.}
\setlength{\tabcolsep}{8pt}
\renewcommand{\arraystretch}{0.95}
\small
\begin{tabular}{c|c|cccc|c}
\toprule
\textbf{Kernel} & \textbf{Standard} & \textbf{SemSeg} & \textbf{Human Parts} & \textbf{Saliency} & \textbf{Normals} &  \multirow{2}{*}{\textbf{$\Delta m$ (\%)}} \\
\textbf{Size $k$} & \textbf{Deviation $\sigma$} & (mIoU ↑) & (mIoU ↑) & (mIoU ↑) & (rmse ↓) &  \\
\midrule
\multicolumn{7}{l}{\textbf{Sweep over kernel size $K$} (fixed $\sigma = 1.0$)} \\
\midrule
$K = 5$ & $\sigma = 1.0$ & 71.20 & 61.37 & 66.16 & 16.15 & +5.32 \\
$K = 7$ & $\sigma = 1.0$ & \textbf{71.25} & 61.38 & \textbf{66.24} & 16.14 & \textbf{+5.39} \\
$K = 9$ & $\sigma = 1.0$ & 71.06 & \textbf{61.46} & 66.07 & \textbf{16.11} & +5.32 \\
\midrule
\multicolumn{7}{l}{\textbf{Sweep over $\sigma$} (fixed $K = 7$)} \\
\midrule
$K = 7$ & $\sigma = 0.5$ & 70.91 & 61.26 & 65.58 & 16.17 & +4.90 \\
$K = 7$ & $\sigma = 1.0$ & \textbf{71.25} & \textbf{61.38} & 66.24 & 16.14 & \textbf{+5.39} \\
$K = 7$ & $\sigma = 1.5$ & 71.05 & 61.31 & \textbf{66.40} & \textbf{16.13} & +5.36\\
\bottomrule
\end{tabular}
\label{tab:gaussian_sweep}
\end{table*}

We analyze the sensitivity of Free Sinewich to the hyperparameters of the
Gaussian low-pass filter applied after sine modulation.
Specifically, we perform one-dimensional sweeps over the kernel size $K$
and the standard deviation $\sigma$, and report the results in
Table~\ref{tab:gaussian_sweep}.

We first vary the kernel size while fixing $\sigma = 1.0$.
As shown in the upper block of Table~\ref{tab:gaussian_sweep},
the performance remains stable across different kernel sizes.
Among the tested configurations, $K=7$ achieves the best overall result,
yielding the highest $\Delta m$ (+5.39).
Both smaller ($K=5$) and larger ($K=9$) kernels produce comparable
performance, indicating that the method is not sensitive to the exact
spatial extent of the filter.

We then fix the kernel size to $K=7$ and sweep the standard deviation $\sigma$.
The results show that the original setting $\sigma=1.0$ consistently
achieves the best trade-off across all tasks.
When $\sigma=1.5$, the performance remains competitive, demonstrating
robustness to moderate over-smoothing.
However, when $\sigma=0.5$, we observe a noticeable drop in performance,
with $\Delta m$ decreasing from +5.39 to +4.90.
We attribute this degradation to insufficient suppression of high-frequency
artifacts introduced by the sine transformation.
With a small $\sigma$, the Gaussian filter becomes too narrow to effectively
smooth oscillatory noise, which adversely affects feature stability and
task-specific modulation.
These results show that Free Sinewich is robust to a wide range of
Gaussian filter hyperparameters, while the default setting ($K=7$, $\sigma=1.0$)
provides the most consistent and optimal performance.

\begin{figure*}[t]
    \centering
    \includegraphics[width=0.9\linewidth]{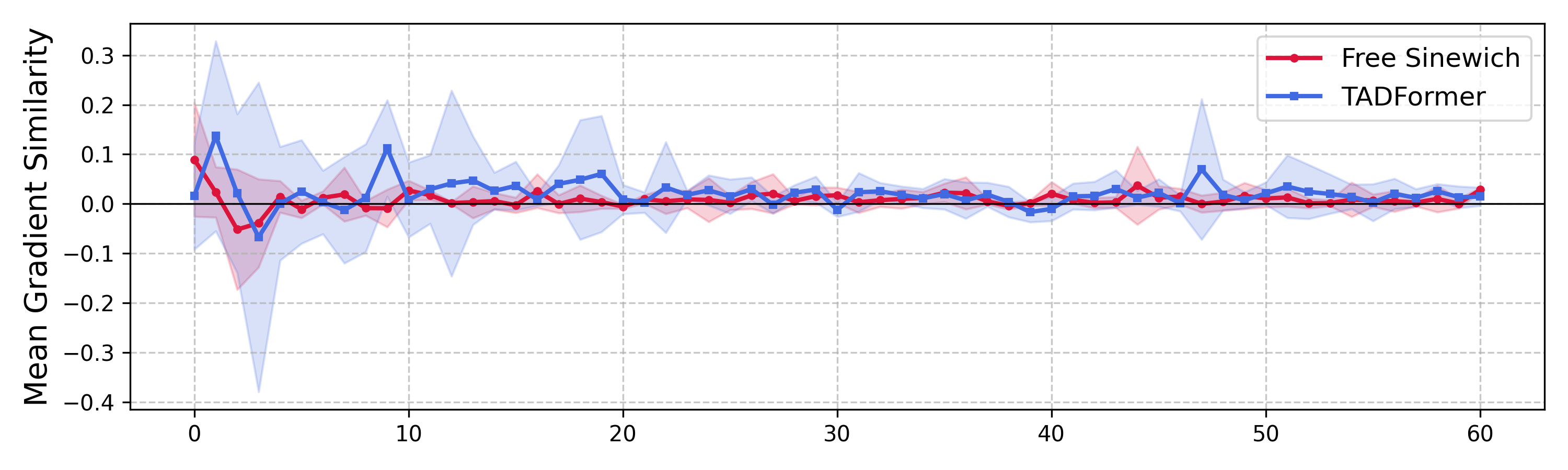}
    \caption{
    Pairwise gradient cosine similarity measured on the shared LoRA base matrix across training epochs (x-axis). 
    Our method maintains near-orthogonal inter-task gradients with reduced variance 
    and lower conflict rate compared to TADFormer.
    }
    \label{fig:grad_sim}
\end{figure*}

\section{Gradient Cosine Similarity Analysis}

To better understand task interference, we measure the similarity between task gradients computed on the shared LoRA base matrices of the encoder. 
For a pair of tasks $i$ and $j$, let $g_i$ and $g_j$ denote the flattened gradients of the shared parameters. 
Their cosine similarity is computed as
\begin{equation}
\mathrm{sim}(i,j) =
\frac{g_i^\top g_j}{\|g_i\|_2 \, \|g_j\|_2}.
\end{equation}

During training, gradients from different tasks are collected at each iteration. 
The similarity values are accumulated across iterations and averaged within each epoch:
\begin{equation}
\bar{s} = \frac{1}{N} \sum_{k=1}^{N} \mathrm{sim}_k(i,j),
\end{equation}
where $N$ is the number of gradient samples in the epoch. 
We also report the variance of these values to quantify gradient stability.

Fig.~\ref{fig:grad_sim} shows a comparison of gradient cosine similarity between TADFormer and Free Sinewich. Compared to TADFormer, our method exhibits lower variance and more stable, near-orthogonal inter-task gradients (similarity $\approx 0$), indicating reduced interference despite shared parameters. 
Although we do not explicitly design a gradient optimization component, the proposed \emph{frequency-switching} mechanism appears to implicitly induce task-specific subspaces within the shared base matrix. 
Combining our method with explicit gradient-based approaches~\cite{navon2022multi,gradient_surgery} remains a promising direction for future work.

\begin{table}[t]
\centering
\caption{\textbf{Results on the Cityscapes dataset.}
All methods use a Swin-Tiny backbone~\cite{liu2021Swin} pretrained on ImageNet-22K~\cite{imagenet} with an HRNet decoder.
MTL-Dec denotes training the decoder only, while MTL-Full denotes full fine-tuning. 
$\Delta m$ denotes the average relative improvement over the single-task baseline.}
\setlength{\tabcolsep}{6pt} 
\renewcommand{\arraystretch}{0.95} 
\small
\begin{tabular}{l|cc|c|c}
\toprule
\multirow{2}{*}{\textbf{Method}} & \textbf{SemSeg} & \textbf{Depth} & \multirow{2}{*}{\textbf{$\Delta m$ (\%)}} & {\textbf{Trainable}} \\
 & (mIoU ↑) & (rmse ↓) &  & \textbf{Param. (M)} \\
\midrule
Single Task & 63.98 & 5.91 & 0 & 56.00 \\
MTL - Dec & 53.30 & 8.54 & -30.59 & 0.97 \\
MTL - Full & 61.03 & 5.76 & -1.03 & 28.49 \\
\midrule
DiTASK~\cite{mantri2025ditask} & 56.08 & 6.35 & -9.89 & 2.58 \\
MTLoRA~\cite{agiza2024mtlora} & 60.88 & 5.84 & -1.83 & 7.23 \\
TADFormer~\cite{baek2025tadformer} & 62.76 & 5.39 & +3.44 & 5.97 \\
\midrule
Free Sinewich & 62.62 & 5.14 & +5.45 & 5.92 \\
\bottomrule
\end{tabular}
\label{tab:cityscapes_results}
\end{table}




\section{Results on the Cityscapes Dataset}

We further evaluate Free Sinewich on the Cityscapes~\cite{Cityscapes} dataset to assess its generalization to large-scale urban scene understanding. 
Following the experimental protocols of TaskPrompter~\cite{ye2023taskprompter} and DiffusionMTL~\cite{diffusionmtl}, 
we consider a two-task setting consisting of semantic segmentation and monocular depth estimation. 
All methods are trained and evaluated under identical settings to ensure a fair comparison.

The quantitative results are summarized in Table~\ref{tab:cityscapes_results}. 
Among parameter-efficient multi-task learning methods, DiTASK and MTLoRA show limited improvements and remain below the single-task baseline. 
TADFormer improves upon these baselines, achieving a positive $\Delta m$ of +3.44 with 5.97M trainable parameters. 
In contrast, Free Sinewich achieves the best overall performance, attaining a $\Delta m$ of +5.45 with only 5.92M trainable parameters. 
These results indicate that Free Sinewich effectively mitigates task interference while maintaining strong parameter efficiency in large-scale outdoor scenes.

\end{document}